\documentclass[conference]{IEEEtran}
\IEEEoverridecommandlockouts
\usepackage{cite}
\usepackage{amsmath,amssymb,amsfonts}
\usepackage{algorithm, algorithmic}
\usepackage{graphicx}
\usepackage{textcomp}
\usepackage{xcolor}
\usepackage{hyperref}

\newcommand{\GPUs}{\text{GPUs}}

\def\BibTeX{{\rm B\kern-.05em{\sc i\kern-.025em b}\kern-.08em
    T\kern-.1667em\lower.7ex\hbox{E}\kern-.125emX}}
\begin{document}
\bstctlcite{IEEEexample:BSTcontrol}

\title{Digital Twin Brain: a simulation and assimilation platform for whole human brain*\\
\thanks{This work is jointly supported by Shanghai Municipal Science and Technology Major Project (No. 2018SHZDZX01), ZJ Lab, and Shanghai Center for Brain Science and Brain-Inspired Technology, and the National Key R\&D Program of China (No. 2019YFA0709502). Corresponding author: Jianfeng Feng Email: jffeng@fudan.edu.cn.
 }
}

\author{\IEEEauthorblockN{Wenlian Lu, Longbin Zeng}
\IEEEauthorblockA{\textit{ISTBI and SCMS} \\
\textit{Fudan University}\\
Shanghai, China \\
\{wenlian,19110850005\}@fudan.edu.cn}
\and
\IEEEauthorblockN{Xin Du}
\IEEEauthorblockA{\textit{School of Computer Sciences} \\
\textit{Fudan University}\\
Shanghai, China \\
xdu20@fudan.edu.cn}
\and
\IEEEauthorblockN{Wenyong Zhang}
\IEEEauthorblockA{\textit{School of Mathematical Sciences} \\
\textit{Fudan University}\\
Shanghai, China \\
17110180029@fudan.edu.cn}
\and
\IEEEauthorblockN{Shitong Xiang, Huarui Wang, Jiexiang Wang, Mingda Ji, Yubo Hou}
\IEEEauthorblockA{\textit{ISTBI} \\
\textit{Fudan University}\\
Shanghai, China\\
\{stxiang19, wanghuarui, 17307110165, mdji, 22110180017\}@fudan.edu.cn}
\and
\IEEEauthorblockN{Minglong Wang, Yuhao Liu, Zhongyu Chen}
\IEEEauthorblockA{\textit{School of Computer Sciences} \\
\textit{Fudan University}\\
Shanghai, China \\
\{21210240337, yhliu20, zhongyuchen20\}@fudan.edu.cn}
\and
\IEEEauthorblockN{Qibao Zheng, Ningsheng Xu, Jianfeng Feng}
\IEEEauthorblockA{\textit{ISTBI} \\
\textit{Fudan University}\\
Shanghai, China \\
\{zhengqb, xns, jffeng\}@fudan.edu.cn}
}

\maketitle

\begin{abstract}
In this work, we present a computing platform named digital twin brain (DTB) that can simulate spiking neuronal networks of the whole human brain scale and more importantly, a personalized biological brain structure. In comparison to most brain simulations with a homogeneous global structure, we highlight that the sparseness, couplingness and heterogeneity in the sMRI, DTI and PET data of the brain has an essential impact on the efficiency of brain simulation, which is proved from the scaling experiments that the DTB of human-brain simulation is communication-intensive and memory-access-intensive computing systems rather than computation-intensive. We utilize a number of optimization techniques to balance and integrate the computation loads and communication traffics from the heterogeneous biological structure to the general GPU-based HPC and achieve leading simulation performance for the whole human brain-scaled spiking neuronal networks. On the other hand, the biological structure, equipped with a mesoscopic data assimilation, enables the DTB to investigate brain cognitive function by a reverse-engineering method, which is demonstrated  by a  digital experiment of visual evaluation on the DTB. Furthermore, we believe that the developing DTB will be a promising powerful platform for a large of research orients including brain-inspired intelligence, rain disease medicine and brain-machine interface.
\end{abstract}


\section{Justification for ACM Gordon Bell Prize}
We have implemented a digital twin brain (DTB) platform to simulate the whole human brain simulations, up to 86 billion neurons and 47.8 trillion synapses, and achieved  the time-to-solution performance for 1 sec of biological time complete within 65 secs, 78.8 secs and 118.8 secs in the wall time for around 7 Hz (alpha rhythm), 15 Hz (beta rhythm) and 30 Hz (gamma rhythm) of average firing rates respectively. We highlight that the heterogeneity of biological structure shows unfriendly to perform high-speed simulation on general HPC but is possibly essential for brain functions. To demonstrate this, a novel data assimilation method was developed  in the DTB platform  and a Pearson correlation over 0.65 between the DTB and its biological counterpart is observed in real world visual evaluation tasks.

\section{Performance Attributes}
\begin{table}[!hp]
\begin{center}
\begin{tabular}{ll}
\hline
Performance Attributes & Our submission\\
\hline
Category of achievement & Scalability, time-to-solution\\
Type of method used & Explicit\\
Results reported on the basis of & Whole application except I/O\\
Precision reported & Mixed precision\\
System scale & Measured on full-scale system\\
Measurement mechanism & Manual measure\\
\hline
\end{tabular}
\label{tab: performance attributes}
\end{center}
\end{table}

\section{Overview of the problem}
\subsection{Human-brain scale simulation}
One of the most ambitious scientific endeavors addressed by neuroscientists ever undertaken is simulating the human brain, aiming to understand how the brain works. The main challenges in simulation consist of the following aspects: large-scale data-based biological structure, simulation speed, and brain functions\cite{Makin2019}. The adult human brain contains one hundred billion neurons and one thousand trillion synapses, of which the cerebral cortex and cerebellum contain 99\%\cite{HerculanoHouzel2009}. Moreover, hundreds of functional regions exist in the brain, and the types of neurons and synapses and the number of connections per neuron differ across regions.

Besides the extremely large scale, the major complexity of the brain synaptic connection structure comprises several other characteristics that prevent the high-efficient simulation. In the view of the sMRI (Structural MRI), DTI (Diffusion Tensor Imaging), and PET (Positron Emission Computed Tomography) data used in this work, based on the resolution of voxels ($3\times 3\times 3~mm^3$), the connections between voxels are of sparseness, with a ratio around 7.2\textperthousand ratio over complete connections. They also show significant couplings, in view of basic units, voxels measured by the structural MRI, of which the mean ratios of inter-voxel over all connections is around 2/7, 6/25, 14/25, and 16/125 for cortex, sub-cortex, brainstem, and cerebellum regions respectively. Finally, they contain high heterogeneity. The ratio of maximum out-connection numbers of each voxel over the minimum is up to 35. These characteristics cause the bottleneck of the simulation scale and speed of the real biological human brain to lie not in the computation but in the bandwidths of memory access and communication, as well as the memory sizes. Furthermore, extreme unbalances of the synaptic connections, which lead to unbalances of the computation, in particular, the unbalances of traffic between computing nodes, are intrinsically against the network architecture of the existing general computing systems.

The computational performance of supercomputers has increased exponentially since their inception and will soon reach 1 exa FLOPS in the 2020s, that is, the 10th to the 20th power of floating point operations per second. This computational power will allow us capable of simulating a human-scale brain model with realistic anatomical and physiological parameters. However, due to the complexity and difficulty mentioned above, no present technology can run large-scale simulations faster than in real-time, typically more slowly, for near whole human brain simulation with real-world biological structure.

The best-known brain simulation project is Blue Brain Project(BBP) launched on 1 July 2005, which uses an IBM `Blue Gene' supercomputer to simulate the neural signaling of a cortical column of the rat brain at an ion-channel level of detail\cite{Markram2006}. The project focused on simulating a single neocortical column, based mainly on data gathered from 15,000 experiments from Markram's and other labs regarding the rat's somatosensory cortex. Based on BBP, the Human Brain Project (HBP) is a multi-national European brain research launched in 2013, aiming to employ highly advanced methods from computing, neuroinformatics, and artificial intelligence to carry out cutting-edge brain research and gain an in-depth understanding of the complex structure and function of the human brain\cite{Amunts2016}. In the HBP, the simulation tools at the whole brain level are developed around the framework of The Virtual Brain (TVB), which focuses on using the mesoscopic laws governing the behaviors of neural populations and uncovering the laws driving the processes of the macroscopic brain network scale\cite{SanzLeon2013}.

In this paper, we focus on implementing simulation and data assimilation of the leaky integrate-and-fire (LIF) neuronal network model of four types of synapses (AMPA, NMDA, GABAa and GABAb) with voxel-wise sMRI-DTI-PET-based biological structure, in addition with cortical micro-column structure, up to 86 billion neurons and 47.8 trillion synapses on the GPU (graphic processing units)-based high-performance computers (HPC) cluster, named digital twin brain (DTB). The GPUs are efficient in computing the differential equations of LIF neurons in parallel, which can speed up the numerical simulation drastically, by using multithread to parallel computation and communication. We utilized an optimization algorithm to balance the load and traffic in and between nodes.

\subsection{Neuron network model}
The whole brain neuronal network model presents the computational basis of the DTB and is composed of two components: the basic computing units and the network structure. The neuronal model we study here is the leakage integrate-and-fire (LIF) neurons equipped with synapses with an instantaneous jump and exponential decay.  Each neuron model receives the postsynaptic currents as the input and describes the generating scheme of the time points of the action potentials as the output. The dynamics of sub-threshold membrane potential $ V_{i} $ for excitatory (inhibitory) neuron can be described as
\begin{equation}\label{eq:cv_equation}
	\left\{
	\begin{aligned}
		&C_{i} \dot{V_{i}}  = -g_{L, i}(V_{i} - E_{L}) + I_{sum}, \\
		&I_{sum}  = \sum_{u}j_{i,u}(V_{i} - E_{u})g_{i, u} + I_{ext}, \\
		&\frac{d j_{i,u}}{dt} =-\frac{1}{\tau_{u}}j_{i, u} + \omega_{u} \sum_{m}\delta_{t - t_{m}^{k}},
	\end{aligned}
	\right.
\end{equation}
where $u\in \{\rm AMPA, NMDA, GABA_{a}, GABA_{b}\}$ denotes different synapse transmitter and $g_{L}, g_{u}$ are the leak, synaptic conductance with equilibrium potentials $ V_{L} $ and $ V_{u} $ respectively. Herein, $-1 / \tau_{u}$ represents the synaptic delay influence, where the AMPA type has a rapid decaying constant and the NMDA channel is much slower. The dimensionless parameters $\omega_{u}$ are the synaptic weights and the gating variables $j_{i, u}$ represent the fractions of open channels of neurons. $m$ belongs to the set of $i$'s presynaptic neuron. $ C_{i} $ is the neuronal capacitance of neuron $i$;  When the membrane potential reaches the firing threshold $ V_{th} $, the neuron emits a spike and the membrane potential is reset at $V_{reset}$ and stays at that value for a refractory period $T_{ref}$. In addition, each neuron in the network receives an external background input $I_{ou}$ to maintain network activity. More specifically, the external currents are modeled as uncorrelated Ornstein–Uhlenbeck processes.

To model the synaptic interactions between neurons in our network, we propose a hierarchical random graph with constraints and multiple edges (HRGCME) due to the limited spatial resolution of brain structural data. Please see article\cite{https://doi.org/10.48550/arxiv.2211.15963} for the details. Each voxel in the brain corresponds to a spatial resolution of $3\times 3\times 3 ~mm^3$, with LIF neurons assigned to each node based on T1-weighted data of sMRI. We use diffusion tensor imaging data (DTI) to determine the connection probability between each voxel. We assume that long-range connections between neuronal populations are only excitatory since inhibitory connections tend to be local. The whole brain is divided into four regions: cortex, sub-cortex, brainstem, and cerebellum. In our experiment, we set the average number of input synapses per neuron in cortex and subcortex voxels as 1000, and that of the brainstem and cerebellum parts as 100. The cortex voxel is modeled as a micro-column structure based on a detailed micro-circuit \cite{Binzegger2009}, while the voxels belonging to the remaining three regions are divided into two parts: excitatory neurons (80\%) and inhibitory neurons (20\%), each with their specific external connection ratio of 6/25, 14/25, and 16/125, respectively, according to the PET data \cite{Finnema2016, Hansen2022}. We further refine the DTI network by incorporating the internal structure of each voxel. See Tab.\ref{tab:example} for the statistical information details. We highlight that the presented network model based on personalized brain structure data and knowledge is quite flexible with giving the number of neurons and input synapse number per neuron possibly for each brain region of interests, according to the computing budget.
\begin{table*}[htbp]
\centering
\caption{Statistical information in the brain network}
\begin{tabular}{|c|c|c|c|c|}
\hline
Regions &  Cortex & Sub-cortex & Brainstem & Cerebellum\\
\hline
Neurons per voxel (mean $\pm$ std) & $834369\pm 127903$& $290432\pm 97269$ & $110796 \pm 46885$& $42528898\pm 8384971$\\
Input connections per neuron & 1000 & 1000 & 100 & 100 \\
Output connections per neuron (mean $\pm$ std) & $1004 \pm 351 $& $952\pm 260$& $202 \pm 185$& $103\pm 34$\\
Maximum output connection per neuron & 5017 & 3612 & 1542 & 833\\
Minimum output connection per neuron & 714 & 760 &  44 &  87 \\
voxel-voxel connection(mean $\pm$ std)& $170\pm 213$ & $250\pm 283$& $147 \pm 109$& $108\pm 105$\\
\hline
\end{tabular}
\label{tab:example}
\end{table*}

\section{Current state of art}
\subsection{Cerebral cortex simulation}
Cortical simulations have a rich history dating back to two classic papers \cite{Farley1954, Rochester1956}.  For more details, please refer to \cite{Brette2007} for an extensive review and comparison of  cortical simulators (NEURON, GENESIS, NEST, NCS, CSIM, XPPAUT, SPLIT, and Mvaspike).

IBM’s Almaden Research Center designed a massively parallel cortical simulator, C2, to run on distributed memory multiprocessors. The C2 simulator implemented a mouse-scale cortical model with 8 million neurons (80\% excitatory) and 6300 synapses per neuron on a 4096-processor BlueGene/L supercomputer\cite{Gara2005}. At a 1 msec resolution and an average firing rate of 1 Hz, the C2 simulator ran 1s of model time in 10s of real-time\cite{Frye2007IBMRR}. When the brain scale is up to 16 million neurons and 8000 synapses per neuron, the simulator could simulate 1s of model time in 33.6 sec of real-time at a mean firing rate of 4.95 Hz\cite{Ananthanarayanan2007b}. Meanwhile, for a rat-scale brain with 57.76 million neurons and 8000 synapses per neuron, the simulator could simulate 1s of model time in 65s at 7.2 Hz average neuronal firing rate\cite{Ananthanarayanan2007}.

However, the synaptic connections of the cortical model introduced above were randomly linked between neurons in different computing processors until in 2008, Izhikevich and Edelman gave a large-scale model of the mammalian thalamocortical system that includes multiple cortical regions, corticocortical connections, and synaptic plasticity\cite{Izhikevich2008}. The shape and connectivity of the model were determined by structural data from three species, human, cat, and rat. In 2009, IBM Almaden Research Center significantly enriched the cortical model with neurobiological data and simultaneously enhanced C2 with algorithmic optimizations and usability features. By using a supercomputer with 147456 CPUs and 144 TB of total memory, the C2 simulator modeled a cat visual cortex with 1.617 billion neurons and 0.887 trillion synapses, roughly 643 times slower than real-time per Hz of average neuronal firing rate \cite{Ananthanarayanan2009}.

One of the simulators for spiking neural network models in EBRAIN is NEST \cite{Gewaltig2007} which enables the user to combine different synaptic dynamics in the same simulation. In 2012, for a network of 20 million neurons with an average firing rate 6.9Hz, the NEST simulation time for 1 sec of biological time was below 240 sec on JUGENE (a BlueGene/P distributed-memory supercomputer system developed by IBM that included 73728 compute nodes), and less than 600 sec for the network of 64 million neurons on the K-computer\cite{Helias2012}. The K computer consists of 88,128 compute nodes, and a compute node in the K system is mainly composed of a CPU, memory modules of 16 GB and a chip for the interconnecting of nodes. The compute nodes are connected with the `Tofu'(torus-connected full connection) interconnect network, a six-dimensional mesh/torus network \cite{Ajima2009TofuA6}. Employing 82944 nodes of the K-computer simultaneously, the network with 1.86 billion neurons and 11.1 trillion synapses took 2481.66 sec to simulate 1 sec of biological time \cite{Kunkel2014}. Using 63504 computer nodes on the K-computer, a larger-scale network with 6.04 billion neurons and 24.5 trillion synaptic connections was simulated at a 0.1 msec resolution and an average firing rate of 4 Hz. The simulation took 322 sec for the simulation of 1 sec biological time \cite{Igarashi2019}.

\subsection{Cerebellum simulation}
The cerebellum, which contains 80\% (69 billion) of all brain neurons and 85 synapses per neuron is known for its regular and repeated crystallized anatomical structure.

RIKEN Center for Computational Science built a cerebellar model with 1048576 neurons, used 4 GPUs to simulate cerebellar activity for 1 sec with a temporal resolution of 1 ms, and achieved real-time simulation \cite{Gosui2016}. On Shoubu, an energy-efficient supercomputer with 1280 computing processors, a cat-scale cerebellar model with 1 billion neurons also implemented real-time simulation in 2019 \cite{Yamazaki2017}. Under the Post-K project, MONET (Millefeuille-like Organization NEural neTwork), a dedicated neural network simulation software, was built in 2019 and designed to calculate layered sheet types of neural networks with parallelization by tile partitioning. Implemented on all 82944 nodes of the K computer, MONET succeeded in simulating a human-scale cerebellar model with 68 billion neurons\cite{Yamaura2020}.

\subsection{Simulation hardware}
Another direction of large-scale simulation is to build special-purpose hardware to emulate spiking neural networks in real-time with ultra-low power consumption as low as the human brain. Due to the fact that modern computers are based on the concept of stored programs introduced by John Von Neumann, rather than optimized for providing brain-like functions, kinds of neuromorphic processors such as SpiNNaker
, BrainScaleS
, TrueNorth
, and Loihi 
have been developed one after another. Compass, a multi-threaded, massively parallel functional simulator and a parallel compiler has been developed by IBM Almaden Research Center since 2011 to fit TrueNorth, which is a compact, ultra-low power, modular architecture consisting of an interconnected and communicating network of extremely large numbers of neurosynaptic cores. On the IBM Blue Gene/Q supercomputer, Compass simulated 256 million TrueNorth cores containing 65 billion neurons and 16 trillion synapses, comparable to the number of synapses in the macaque monkey cortex. At an average neuron spiking rate of 8.1 Hz, the simulation is 388 slower than real-time \cite{6468524}. See Tab.\ref{tab:Survey} for the brain simulation details.
\begin{table*}[htbp]
    \centering
    \caption{Survey of large-scale brain simulation}
    \begin{tabular}{c|c|c|c|c|c|l}
        Brain simulation & Neurons & Synapses & Firing rate & Time-to-solution & Temporal resolution & Connection topology \\\hline
        Frye\cite{Frye2007IBMRR}& 8 million & 50.4 billion & 1 Hz & 10 & 1 msec & completely random \\
        Ananthanarayanan\cite{Ananthanarayanan2007}& 16 million & 128 billion & 4.95 Hz & 33.6 & 1 msec & completely random\\
        Ananthanarayanan\cite{Ananthanarayanan2007b}& 57.67 million & 461 billion & 7.2 Hz & 65 & 1 msec & completely random\\
        Ananthanarayanan\cite{Ananthanarayanan2009}& 1.617 billion & 8.87 trillion & 19.1 Hz & 643 per Hz & 0.1 msec & based on cats anatomical data\\
        Helias\cite{Helias2012}& 20 million & 225 billion & 6.9 Hz & $<$ 240 & 0.1 msec & -\\
        Helias\cite{Helias2012}& 64 million & 720 billion & 6.9 Hz & $<$ 600 & 0.1 msec & -\\
        Kunkel\cite{Kunkel2014}& 1.86 billion & 11.1 trillion & - & 2481 & 0.1 msec & -\\
        Igarashi\cite{Igarashi2019}& 6.04 billion & 24.5 trillion & 4 Hz & 322 & 0.1 msec &  based on neurons distances\\
        Gosui\cite{Gosui2016}& 1.048 million & - & 1 Hz & 0.88 & 1 msec & based on cats anatomical data \\
        Yamazaki\cite{Yamazaki2017}& 1 billion & - & - & 0.783 & 1 msec &based on cats anatomical data \\
        Yamazaki\cite{Yamaura2020}& 68 billion & - & - & 416 & 0.1 msec & based on known anatomical data\\
        Preissl\cite{6468524}& 65 billion & 16 trillion & 8.1 Hz & 388 & 1 msec & based on macaque brain\\ \hline
        &&& 7 Hz & 65 && Feng Jianfeng's sMRI-DTI data,\\
        Our work& 86 billion & 47.8 trillion & 15 Hz & 78.8 & 1 msec & PET-based synapse data in literature and \\
        &&& 30 Hz & 118.8 && cats' anatomical cortical microcolumn data\\ \hline
    \end{tabular}
    \label{tab:Survey}
\end{table*}

\section{Innovations realized}
\subsection{Summary of contributions}
\begin{enumerate}
    \item We construct a human-brain-scaled spiking neuronal network with an sMRI-DTI-PET-based biological structure, which is significantly different from the existing works that simulated billions of neurons with a network structure that is simple homogeneous and toy in science with nothing with the human brain.
    \item The biological brain structure is highly sparse, heterogeneous and coupled, which is in fact against the general HPC system which is good at solving balanced and homogeneous computing problems. We present a partitioning algorithm to balance and minimize the traffic between different GPUs, as well as other parallel operations, to enhance efficiency and achieve the leading time-to-solution.
    \item We also present a mesoscopic data assimilation method that can conduct statistical inference of the extremely large-scaled neuronal networks and thus enable the human-brain-scaled model can be assimilated to simulate and emerge dynamics with real cognitive behaviors, rather than the existing works that simulated billions of neurons without any brain functions.
\end{enumerate}

\subsection{Parallel operations of computation and communication}
\begin{figure}
    \centering
    \includegraphics[width=.4\textwidth]{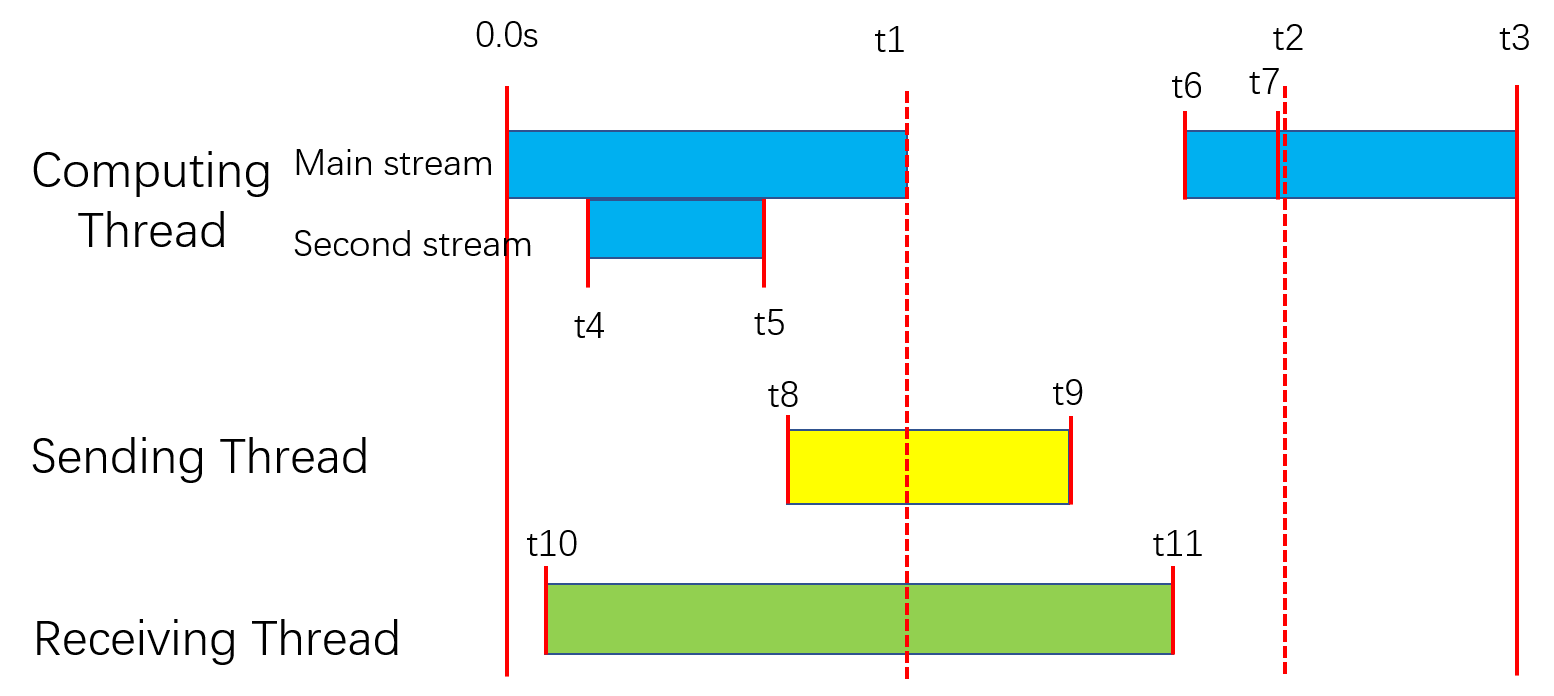}
    \caption{Schematic diagram of parallel operations and time anchor}
    \label{fig:stat}
\end{figure}

The simulation of DTB mainly contains three threads to implement parallel operations of computation and communication, and brain network computation is implemented on GPUs, which are dedicated hardware for parallel computing. Since the brain network model is defined by differential equation\ref{eq:cv_equation}, their state variables can be kept within GPU memory, thus spikes sequence is communicated between GPUs. A detailed flow diagram can be seen in Fig.\ref{fig:stat}. For convenience, we rewrite the second part of the equation \ref{eq:cv_equation} as follows.
\begin{equation*}
    J_{intra} = \sum_{m\in\text{same GPU}} \delta_{t - t_{m}^{k}}, \quad
    J_{inter} = \sum_{m\in\text{other GPU}} \delta_{t - t_{m}^{k}}.
\end{equation*}

For neuron i, $J_{intra}$ and $J_{inter}$ calculate the weighted sum of spikes of presynaptic neurons loaded in the same GPU and other GPUs respectively.

{\em\bf Computing thread} is used to manage GPUs to calculate brain networks. Specifically, GPUs update the neuronal membrane potential based on state variables in the GPUs memory at the beginning, then collect spikes sequence and add an additional stream to send information such as spikes sequence to the CPU, and meanwhile, the original stream continues calculating the $J_{intra}$. After completing the $J_{intra}$ calculation, the thread pauses and waits for the spikes sequences from other GPUs. The earlier the GPU receives information from other GPUs, the earlier the thread starts updating the synaptic current equation. That's why we design parallel operations of computation and communication.

{\em\bf Sending thread} is used to send information to other CPUs. After the spikes sequence is copied from the GPUs to CPU memory, sending thread cuts the sequence into multiple groups and sends it to corresponding CPUs. Each rank sends spike information in the order of rank to some extent preventing communication contention, that is prioritizing the larger rank than itself. When the sending to the largest rank is completed, sending starts at rank 0 and continues until its previous one. This method is similar to the pairwise approach which is a mature technology in collective communication. Since we need to count the average firing rate of different neuronal populations in network model simulation even to collect state variables such as neuronal membrane potential and synaptic current, the slave node also requires sending information to the master node. The sending speed depends on the amount of data sent and whether the receiving node is communicating congestion.

{\em\bf Receiving thread} is used to receive information from other CPUs. stored in the CPU global memory. Because the CPU of other nodes may send information earlier than the CPU where this thread is located, the start time of the receiving thread needs to be as early as possible, slightly later than the computing thread. After the receiving thread has received all information, the computing thread needs to copy information from the CPU to GPU. Then computing thread continues calculating $J_{inter}$ and updating $J$ and $I$ equations.

\subsection{Partitioning algorithm}
The proposed partitioning scheme assigns neurons to GPUs in order to minimize the traffic between different GPUs. Since communications between neurons on the same GPU are much faster than that between neurons across GPUs, this algorithm reduces communication latency by assigning a set of neurons that have strong communication demand between each other on the same GPU. Furthermore, this algorithm makes the traffic between each pair of GPUs in the system as balanced as possible to avoid network congestion thus improving delay performance.

The objective of neurons partition in brain simulations is to assign neurons to GPUs in order to balance the traffic load for inter-GPU communications; that is, minimize the traffic volume of the inter-GPU communication session with the heavies load. Therefore, mapping $n$ neurons to $N$ GPUs in a brain simulation can be formulated as a $N$-way partition for a graph; that is, finding a partition $P=[V_1, ..., V_N]$ that assigns $n$ neurons to $N$ GPUs, with the goal of minimizing the traffic volume on the edge between the pair of GPUs that has the heaviest traffic load. The huge number of neurons with their interconnections in the whole human brain simulation presents a scalability challenge to this partitioning problem. In neuron-level human brain simulations, the $n$ interconnected neurons can be modeled as a weighted directed graph $G=(V, E)$, where the neurons are represented as vertices $V=\{v_1, ...,v_n\}$ and the connection between a pair of neurons $v_i$ and $v_j$ is represented by the edge $e_{i,j} \in E$, $E \subseteq V \times V$. The weight of vertex $v \in V$ is the size of the corresponding neuron and the weight of edge $e_{i,j}$ is the interconnection probability for the pair of neurons $v_i$ and $v_j$. Therefore, we can formulate the neuron partition problem to minimize the following function, where $D_{ij}$ denotes communication traffic from $i$ to $j$:
\begin{equation}
F(P)=\max_j\sum_{i=0}^{N}D_{ij}
\label{equation:formulation}
\end{equation}

The solution $P$ to the neurons partition problem in brain simulations is subject to some constraints. First, the solution $P$ must divide $V$ into $k$ disjoint subsets; that is,
\begin{equation}
\bigcup_{i=1}^{N}{V_i}=V \ \text{and} \ V_i \cap V_j = \phi \ \forall i \neq j.
\label{constraint:1}
\end{equation}

Second, the limited capacity of each GPU requires that the sum of the neurons' sizes simulated by each GPU is lower than a predefined upper bound $\gamma$, i.e.
\begin{equation}
 \sum_j^{v_j\subseteq V_i} (\alpha * s_{1i} + \beta * s_{2i} + \mu * s_{3i} ) \leq \gamma  \label{constraint:2}
\end{equation}

While maintaining the balance of the volume of each GPU, we use a greedy strategy to assign neurons to every GPU to ensure high cohesion among neurons within a GPU and low coupling between neurons on different GPUs. It is worth noting that this greedy assignment is achieved by optimizing traffic between neurons (which is denoted as $D$), and the detailed calculation method of this traffic can be referred to \cite{icpp2022,9785756}. Finally, the algorithm updates the solution and outputs a table for neuron-GPU mapping.

\subsection{Data structure}
\begin{figure}[htbp]
    \centering
    \includegraphics[width=.49\textwidth]{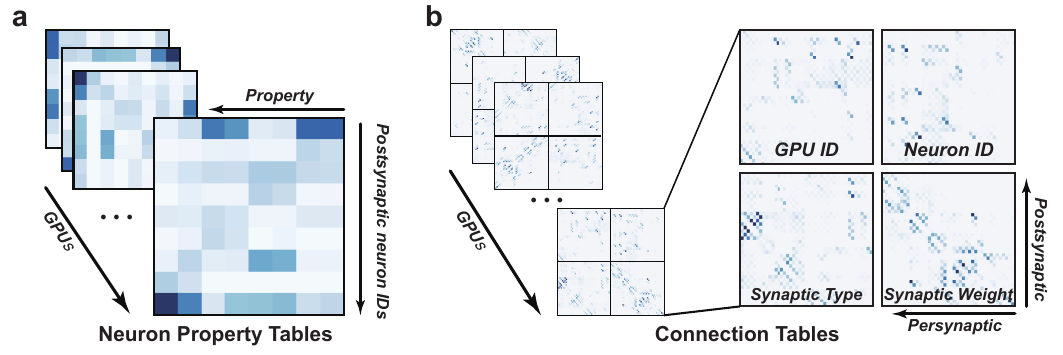}
    \caption{Schematic diagram of neuron property tables(a) and connection tables(b)}
    \label{fig:Data_Tables}
\end{figure}

According to the information mentioned above, we generate neuron property tables and connection tables for brain network simulation, shown in Fig.\ref{fig:Data_Tables}. In neuron property tables, a vertical array represents an array of postsynaptic neuron IDs. Each row element stores the property of the postsynaptic neuron, such as membrane potential equation parameters, characterized by configurable parameters sufficient to meet customized modes. Since we deploy neurons on GPUs and each GPU loads part neurons, thus we divide the neuron property table into multiple groups whose number equals the number of GPUs, that is to say, each GPU corresponds to one neuron property table. Similarly, in connection tables, we generate four connection matrices for each GPU to store the following information: GPU ID where the presynaptic neuron is located, neuron ID where the presynaptic neuron is in its located GPU, synaptic type, and synaptic weight. Considering the sparsity of the brain network, the number of rows of each matrix is equal to the number of neurons located in this GPU and the number of columns is equal to the number of input synapses per neuron. 

If each neuron sends an individual message to each of its target neurons on each communication time resolution, GPUs need to search postsynaptic neuron ID from connection tables for each spike. This is inefficient and may cause communication congestion. In a massively parallel implementation each compute node will house and process several neurons. Hence, each neuron can group messages to all targets that reside on a single compute node into a single message waiting for centralized sending and receiving.

\subsection{Hierarchical mescoscopic data assimilation}
From the simulation of DTB, we can collect the neural activities (i.e., spike raster) and the corresponding firing rates. The time series of the blood-oxygen-level-dependent (BOLD) signal from each voxel are formulated by the Balloon-Windkessel model and generate the simulated BOLD signals\cite{Friston2000}. To fit the experimental BOLD signal with the simulated BOLD signal, we proposed a framework of hierarchical mescoscopic data assimilation (HMDA) to estimate the model parameters, more specifically, the synaptic conductance of each neuron. However, due to the huge amount of parameters (more than 10 trillion parameters for 86B neurons in total), we take each population, according to the definition of the MRI voxel or each layer of the microcolumn structure with excitation/inhibitory neuron types, and assume that the conductance parameters of the same type of the neurons in the same population follow the same distribution, or, equivalently, sampled from the same hyper-parameters. The aim of the data assimilation is to statistically infer the brain-scale neuronal network by tracking the BOLD signals. Refer to \cite{https://doi.org/10.48550/arxiv.2206.02986} for details. The hyper-parameters are estimated and each conductance value is sampled according to the distribution with the hyper-parameters. In the implementation, we estimate a model with the same structure and a lower size of 0.2 billion neurons and 20 billion synapses to achieve the hyper-parameters and sample the parameters of the model of the human brain scale (86 billion neurons and 47.8 trillion synapses) from these hyper-parameters.

In implementation, we employed disjoint sets of nodes to simulate different samples of brain models; then we collected the population-wise spike rates from each node via GRPC to a central node to calulcate voxel-wise hemodynamics as the BOLD signals and conduct iterations of the HDMA to adapt the hyper-parameters; finally we resampled and updated the parameters of neurons from the hyper-parameters on each simulation node. By this way, we realized the assimilation process to fit the BOLD data on the DTB.

\section{How performance was measured}
\subsection{DTB simulator architecture}
We completed simulation experiments of human brain network of 86 billion neurons and 47.8 trillion synapses, corresponding to the full size of the human brain, on a cluster of heterogeneous computing nodes. When simulating, the main steps include the following steps: (1) generate neuron property tables and connection tables, (2) run the simulator to load data and deploy neurons to GPUs, (3) initialize the simulator, (4) run the simulator to complete simulating.

All numerical calculations pertaining to neurons use a forward Euler method with a calculation step of 1 msec. The time of spike communication is also 1 msec, achieved by MPI (message-passing interface) communication cables which can be regarded as axons interconnecting the neurons.

\subsection{Hardware and Software Environment}
The C++ program using C++ compiler (g++) version 7.3.1, was developed with MPI library functions and rocm-4.1.0 hipcc compiler for GPU. The Python program for data assimilation was developed using Python version 3.7.11. The Python and C program runs in a Linux environment.

The cluster in Advanced Computing East China Sub-Center, has 3503 computing nodes, each of which has single  32-core processors operating at 2 GHz and 128 GB of DRAM memory. Each node also has 4 GPUs operating at 1.10 GHz and each GPU has 16GB HBM2 working on 800MHz, with 1TB/s memory bandwidth. GPU communication between each other within one node is through share memory, while communication across nodes is through a 200Gbps Full Duplex Infiniband network. Thus the total theoretical computational performance is 126.26 PFLOPS, and the total amount of GPU memory is 224.192TB.

\subsection{Statistical interface}
Timings of computation and communication are done manually by recording time anchors. We have designed a performance statistics interface to collect time anchors and Flops of each GPU, as shown in Fig.\ref{fig:stat} and Tab.\ref{tab:stat_label}. The entire calculation cycle starts with $V$ equation update at $t_0=0$ and ends with completing $I_{sum}$ update at $t_{3}$. Thus total simulation time is $T_{sim} =t_{3}$. Neurons equation computation time is $T_{com}=t_{3}-t_{2}+t_{1}$ and sending time cost $T_{send} = t_{9}-t_{8}$. Because the time anchor t1 for different GPUs to complete the membrane potential update and spike information copied is different, the start time of sending threads of different GPUs is also different. Receiving thread needs to start as early as possible, so it is sending cost time rather than receiving time cost $T_{rec}=t_{11}-t_{10}$ that represents communication time. Furthermore, we calculate the intra-node communication time $T_{\cdot,intra}$ and inter-node communication time $T_{\cdot, inter}$ of spike information and it is obvious that $T_{send, intra}+T_{send, inter}$= $T_{send}$. Besides, we calculate the communication traffic within and between GPUs, and the FLOPS (floating-point operations per second) where floating-point operations are counted as the sum of addition and multiplication times in equation updates.
\begin{table*}[htbp]
    \centering
    \caption{Statistical indicators and symbol}
    \begin{tabular}{|c|c|c|c|c|c|c|}
    \hline
        Statistical & computing before & computing after & computing & sending copy & sending copy & receiving copy\\
        indicators & communication & communication & end & start & end & start \\ \hline
        symbol & $t_1$ & $t_2$ & $t_3$ &$t_4$ & $t_5$ & $t_6$ \\ \hline
        Statistical & receiving copy & sending  & sending & receiving & receiving & sending duration \\
        indicators & end & start & end & start & end & inter node \\ \hline
        symbol & $t_7$ & $t_8$ & $t_9$ & $t_{10}$ & $t_{11}$ & $T_{send,inter}$ \\ \hline
        Statistical & sending duration & receiving duration & receiving duration & sending byte size & sending byte size & receiving byte size \\
        indicators & intra node & inter node & intra node & inter node & intra node  & inter node \\ \hline
        symbol & $T_{send,intra}$ & $T_{rec,inter}$ & $T_{rec,intra}$ & - & - & - \\ \hline
        Statistical & receiving byte size & FLOPS update  & FLOPS update & FLOPS update & FLOPS update & FLOPS update \\
        indicators & intra node & membrane $V_i$ & inner j presynaptic & outer j presynaptic & j presynaptic & i synaptic \\ \hline
        symbol & - &-&-&-&-&- \\
        \hline
    \end{tabular}
    \label{tab:stat_label}
\end{table*}

\section{Performance result}
\subsection{Human-brain-scale network simulation}
We performed a spiking neuronal network simulation of human-brain scale, using 3503 computer nodes and 14012 GPUs. The whole brain model consists of 86 billion neurons and 47.8 trillion synapses (AMPA, NMDA GABAa and GABAb) with the topological structure mentioned in the neuron model section. In this model, the number of neurons loaded on each GPU is dominated by the GPU memory size. Due to the simulator storing enough neuron properties as configurable parameters to meet custom patterns, the data structure determines that up to 61.3 million neurons are loaded on each GPU in this subsection.

\begin{figure}[htbp]
    \centering
    \includegraphics[width=.49\textwidth]{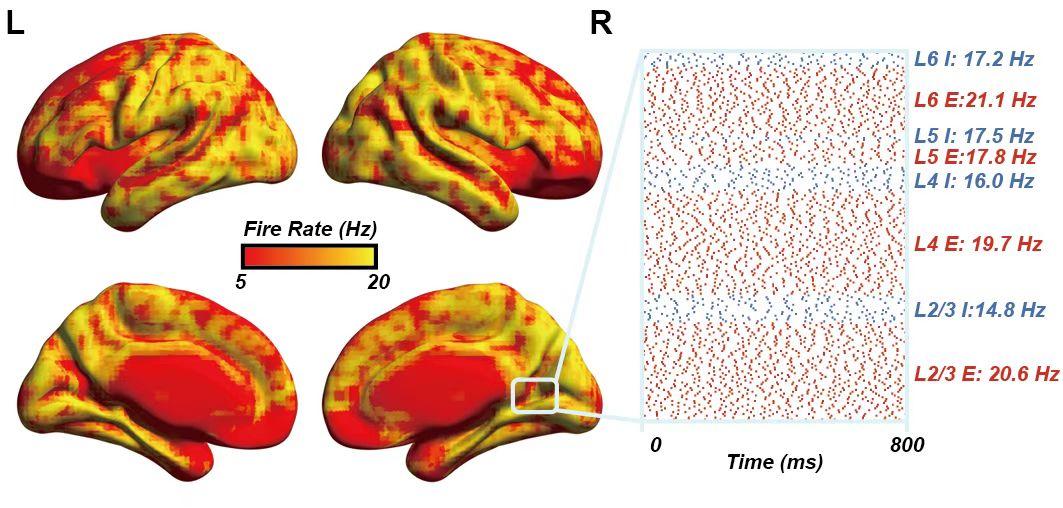}
    \caption{Human-brain-scale network simulation}
    \label{fig: raster}
\end{figure}

Actually, according to the synchronous scheme, the total simulation time cost depends on the GPU of  the slowest computation thread. Thus we collect all-time statistics during 4s simulation by statistical interface and define total simulation time using the last 800 msec (the period of the functional MRI data in our experiment) data as follows.
\begin{equation*}
    T_{label} = \frac{1}{800}\sum_{t=1}^{800} \mathop{max}\limits_{i\in \GPUs} T_{label,i,t}
\end{equation*}
where label as a symbol $\in\{$simulation, computation, sending, receiving$\}$. Time-to-solution is defined as $T_{tos} = T_{sim}$ per second. For the whole brain network with an average firing rate 7Hz, the simulation time for 1 sec of biological time is 65s, which means $T_{tos} = 65$. By balancing the synaptic conductivity $g_{AMPA}$ and $g_{NMDA}$, we adjust the average firing rate of the whole network as 15Hz and 30Hz, whose time-to-solution are 78.8 and 118.8 respectively. Due to limited space, taking 15 Hz as an example, Fig.\ref{fig: raster} shows the firing rate distribution in brain voxels and the spike raster plot of micro-column layers in the cortex voxels.

According to the statistical interface, the following statistical indicators can be calculated in Tab.\ref{tab:FLOPS}, which can represent the utilization rate of hardware in our simulation. There is indeed a big
gap from the actual to the theoretical peak FLOPS and bandwidth, as the sparsity of spikes cannot take advantage of coalesced memory access. Indeed, the bottleneck of the capacity of the number of neurons in the network model following the biological structure lies at the limit of the memory size of each GPU, according to the number of input neurons of the neurons of each GPU multiplied with the coding size of neurons, which can be referred to Sec. V.D for the details.
\begin{table}[hpbt]
    \centering
    \caption{FLOPS and bandwidths at different firing rate}
    \scalebox{0.75}{
    \begin{tabular}{c|c|c|c|c}
        Statistical & FLOPS on  & FLOPS on inner & FLOPS on outer & FLOPS on j \\
        indicators & membrane $V_i$ & j presynaptic & presynaptic & presynaptic  \\ \hline
        At 7 Hz & 54.4 GFLOPS & 2.20 GFLOPS & 2.89 GFLOPS & 72.6 GFLOPS \\
        At 15Hz & 50.8 GFLOPS & 3.72 GFLOPS & 3.03 GFLOPS & 72.7 GFLOPS \\
        At 30Hz & 46.5 GFLOPS & 4.57 GFLOPS & 2.99 GFLOPS & 72.5 GFLOPS \\ \hline
        Statistical & FLOPS on I & Sending bandwidths & Sending bandwidths & Copy sending \\
        indicators & synaptic & intra node & inter nodes & bandwidths\\ \hline
        At 7 Hz & 0.15 TFLOPS & 2.37 GB/s & 0.789 GB/s & 3.09 GB/s\\
        At 15 Hz & 0.15 TFLOPS & 5.00 GB/s & 1.06 GB/s & 3.56 GB/s \\
        At 30 Hz & 0.15 TFLOPS & 7.23 GB/s & 1.38 GB/s & 3.69 GB/s \\
    \end{tabular}
    }
    \label{tab:FLOPS}
\end{table}

\begin{figure}[htbp]
    \centering
    \includegraphics[width=.49\textwidth]{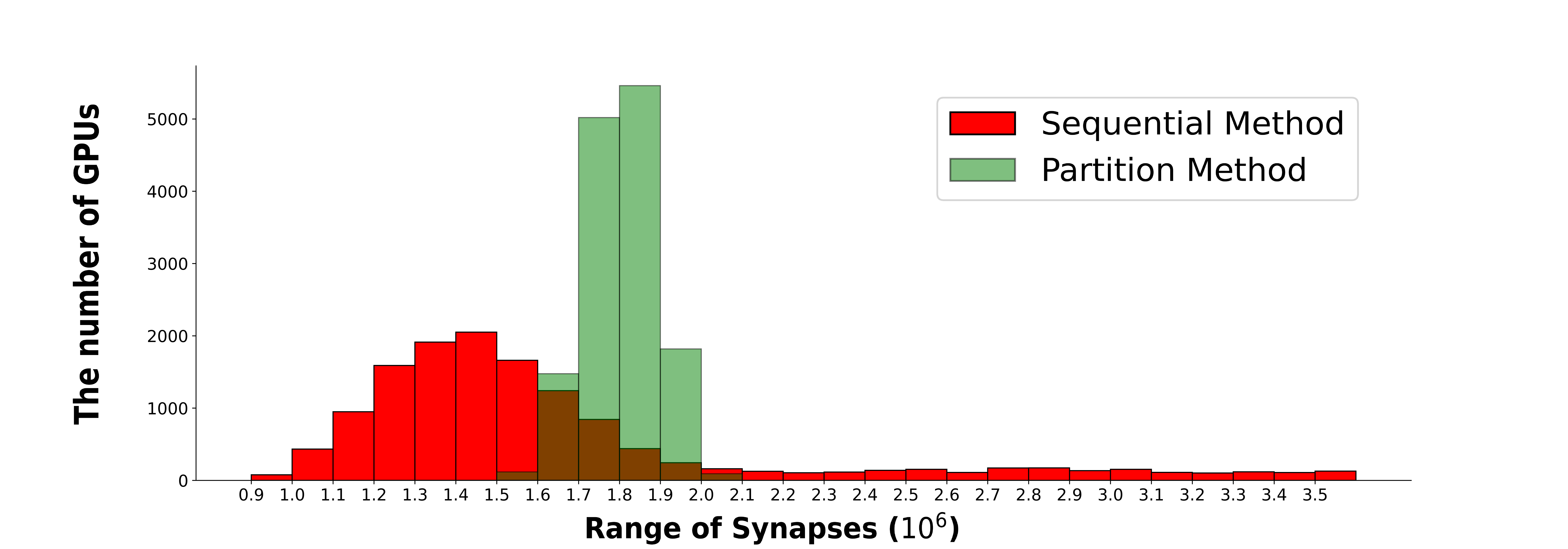}
    \caption{Comparison of inter-GPU synapses simulated with different neuron partitioning methods}
    \label{fig:compared_method}
\end{figure}

To demonstrate the effectiveness of partition algorithms on human-scale networks, a comparison of inter-GPU synapses simulated with different neuron partitioning methods is presented in Fig.\ref{fig:compared_method}, where the red and green columns give the histograms of frequency distribution for the total number of inter-GPU synapses generated in the simulation system using the sequential method and the proposed partition method, respectively. It can be seen from the red column that the sequential method causes significant fluctuation in the amounts of synapses from different GPUs, which may easily cause traffic congestion on some network links in the supercomputer. The green column shows that the proposed partitioning algorithm may greatly reduce the difference in synapse numbers across GPUs. Therefore, the proposed algorithm is more effective in balancing traffic and avoiding congestion compared to the sequential method.

\subsection{Accuracy verification}
To verify the correctness of communication calculation between nodes and the simulation of the neuronal network model on CUDA, we compared the results obtained from the CUDA simulation with those obtained from the CPU simulation. To this end, we generated a large-scale neuronal network of neurons with an embedded independent subgraph and simulated it on CUDA. At the same time, we simulated the embedded independent subgraph on the CPU and compared the numerical results with those obtained from the independent subgraph on CUDA. To be more specific, we randomly sampled a portion of neurons from the independent subgraph and compared their membrane potentials, synaptic currents, and spike times. The computed results showed that the timing (in msec) of spike emission times obtained from the two computation methods were exactly consistent, with relative mean square errors of both membrane potentials and synaptic currents less than $10^{-4}$ during total 1 sec simulation.

\subsection{Different connection topologies}
Here we assess the impact of varying input synaptic numbers per neuron of cortical and subcortical models on computational efficiency. In order to facilitate model scaling and variable control, unless otherwise specified, the large-scale network used in the rest of this section only models the cortex and subcortical regions as part of the whole brain.

\begin{figure}[htbp]
    \centering
    \includegraphics[width=0.24\textwidth]{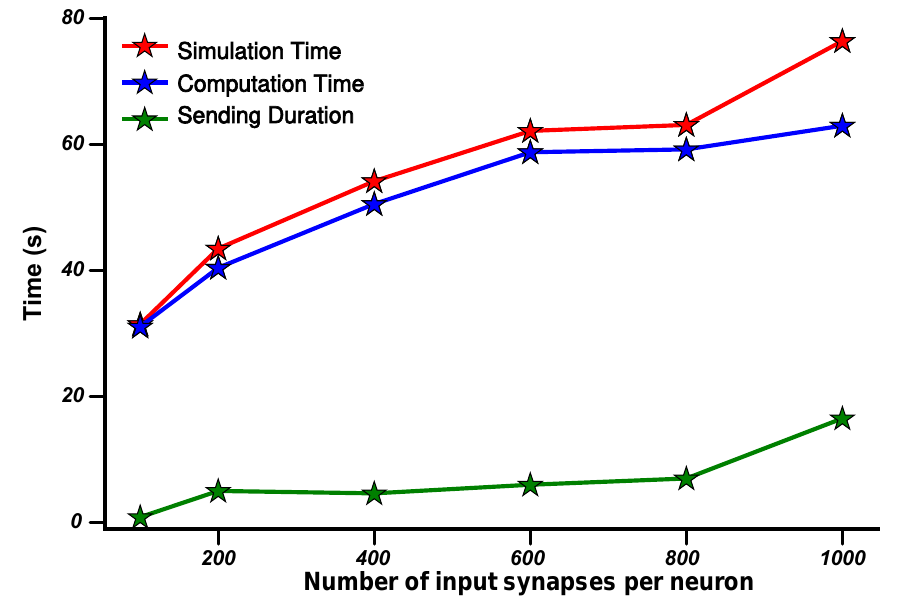}
    \caption{\textbf{Different degree}. We controlled the number of neurons (1.5 billion) and varied the input degrees of the network and the number of GPUs. The horizontal axis is the number of input synapses per neuron, and the vertical axis is the time (s) spent for 1 sec simulation.}
    \label{fig:Ds}
\end{figure}

To save computing resources, we built networks of smaller neuron numbers, i.e., 1.5 billion neurons with varying input synapse numbers per neuron from 100 to 1000. Because the storage of synaptic information also leads to an increase in memory, thus the GPUs used to implement networks also range from 100 to 1000. On one hand, An intuitive impact is an increase in the number of input synapses per neuron leads to an increase in computing resources, and the increase in the number of GPUs also leads to an increase in communication traffic and total communication time. Fig.\ref{fig:Ds} also provides verification. On the other hand, the increase in the number of input synapses per neuron implies that a single neuron would receive more spikes from presynaptic neurons. So the floating point operations of calculating the weighted sum of spikes also increase, resulting in longer computation time. 

According to the trend of the total simulation time curve in Fig.\ref{fig:Ds}, if low computational efficiency can be tolerated, the average degree can reach 5000, or even exceed that of the human brain.

We highlight that one of the biggest differences between our simulator and other simulators such as C2\cite{Ananthanarayanan2007b} is that we have fully utilized DTI data and applied the partition algorithm when determining the connection probability between each voxel. 
Here we designed a kind of connection topology manually, named artificial brain, where each voxel in our model just connects with two neighbor voxels. This topology guarantees efficient communication between GPUs since spike information can be transferred at most to two GPUs. As a comparison, we also generated a brain model that does not use partition algorithms, named the brain model without partition to be distinguished from our brain model. We adjusted the AMPA synaptic conductivity so that the average firing rate of neurons in these networks is all 7.6Hz.
\begin{figure}[htbp]
    \centering
    \includegraphics[width=0.24\textwidth]{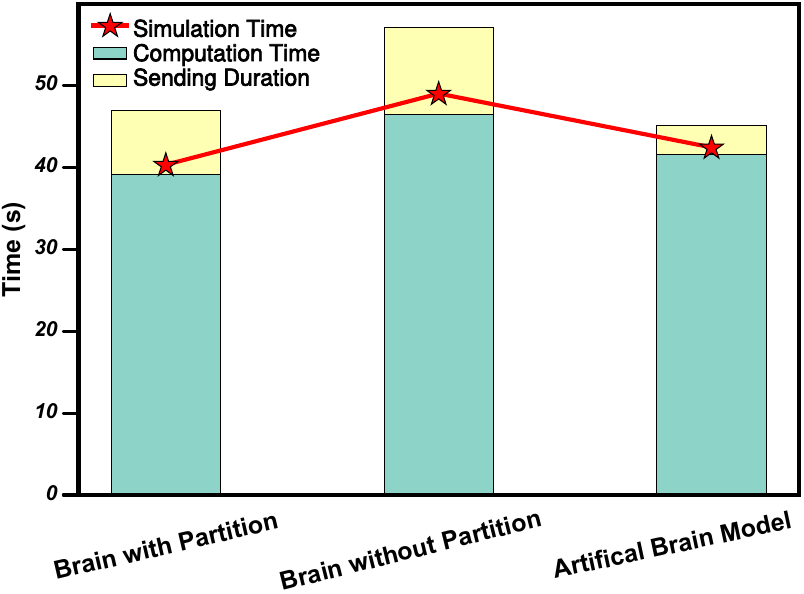}
    \caption{\textbf{Different structure}. We simulate three different structures with the same number of neurons (15 billion), input synapse per neuron (100) and GPUs (1000): brain with partition algorithm, brain without partition algorithm and artificial brain model. The horizontal axis is the different structures of the network, and the vertical axis is the time (s) spent for 1 sec simulation.}
    \label{fig:DifStr}
\end{figure}

As shown in Fig.\ref{fig:DifStr}, with the same network neurons and the same number of input synapses per neuron, simulation on the artificial brain has significantly high efficiency compared with the sequential brain. In other words, if we ignore the biological structure of the human brain, the simulation speed would increase, but this is not conducive to our research on brain mechanisms. But from the comparison between the sequential brain and the DTB standard brain, it can be seen that by partition algorithms, we can improve computational efficiency by balancing computation and communication.

\subsection{Weak scaling}
\begin{figure}[htbp]
    \centering
    \includegraphics[width=0.24\textwidth]{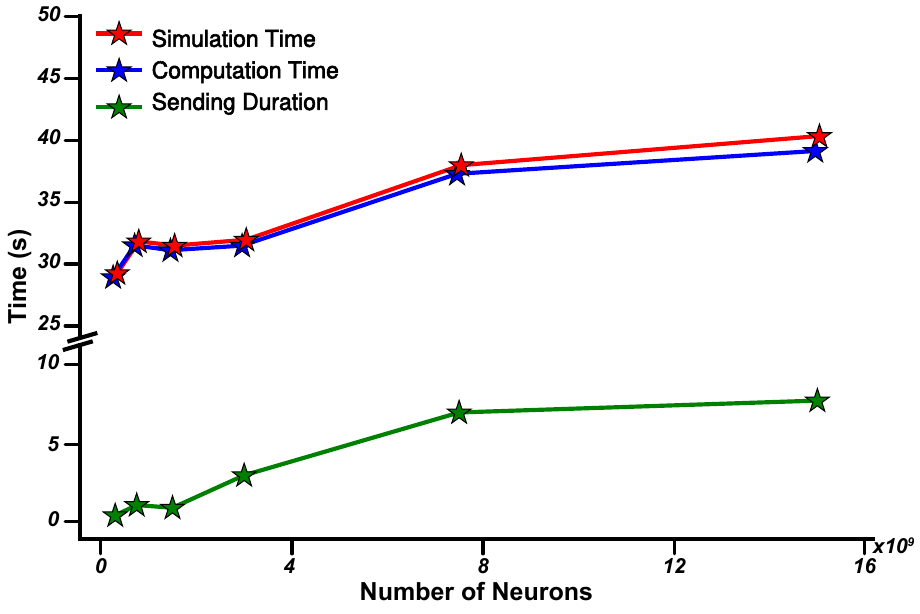}
    \includegraphics[width=0.24\textwidth]{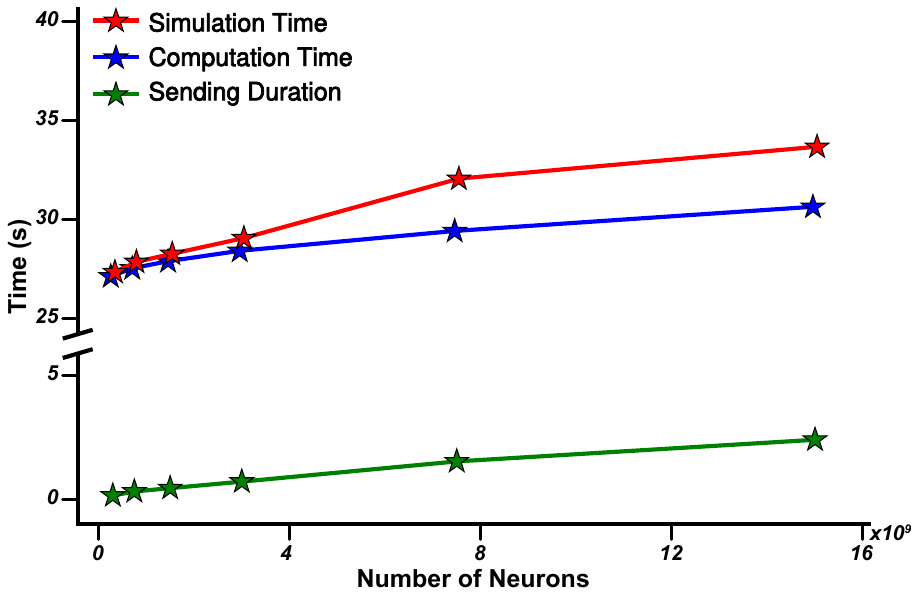}
    \caption{\textbf{Weak scaling property}. We varied the number of neurons: 0.3 billion, 0.75 billion, 1.5 billion, 3 billion, and 15 billion(left). Another time criterion is to represent the average time cost of GPUs(right). The horizontal axis is the number of neurons, and the vertical axis is the time (s) spent for 1 sec simulation.}
    \label{fig:Ws}
\end{figure}

An important advantage of our simulator is to achieve a large number of parallel calculations to update neuron equations which can be distributed to a large number of GPUs. Weak scaling relates to scaling the network model in such a way that the number of neurons loaded on each GPU remains relatively fixed as the number of nodes is increased. Perfect weak scaling yields that the computation time remains unchanged across any network size. In this subsection, the number of neurons on each GPU is fixed as 15 million. By varying the number of GPUs from 20 to 1000, different scales of cerebral cortex models with the same network topology have been simulated at an average firing rate of 7.6 Hz, as shown in Fig. \ref{fig:Ws}.

Due to the increase in the number of GPUs, the same spike information needs to be sent to different GPUs, resulting in an increase in duplicate parts during spike information copied. This in turn leads to an increase in total communication traffic, resulting in an increase in the total communication time cost. Besides, according to Fig.\ref{fig:Ws}, computation time seems to increase as well. We believe that is because the computation time on each GPU is different, and the computation time cost depends on the lowest GPU. Since our brain network is heterogeneous, the partitioning algorithm is applied to each network. Therefore, we cannot guarantee the voxel order holding on each GPU. There may be a communication imbalance between some GPUs. So we define another simulation time $\hat{T}_{label}$ to represent the average time cost between GPUs as follows.
\begin{equation*}
    \hat{T}_{label} = \mathop{mean}\limits_{i \in \GPUs} T_{label, i}
\end{equation*}

In the ideal case of weak scaling, computation time cost remains unchanged, and the time to solution slightly increases due to the increase in communication time, as shown in Fig.\ref{fig:Ws}.

\subsection{Strong scaling}
It is believed that the computational performance of supercomputers has been increasing exponentially. Thus strong scaling is to study the effect on simulation time for a fixed brain scale with increasing computing power, which is represented by the number of GPUs. Perfect strong scaling yields that the computational time halves by doubling the computational power.

\begin{figure}[htbp]
    \centering
    \includegraphics[width=0.24\textwidth]{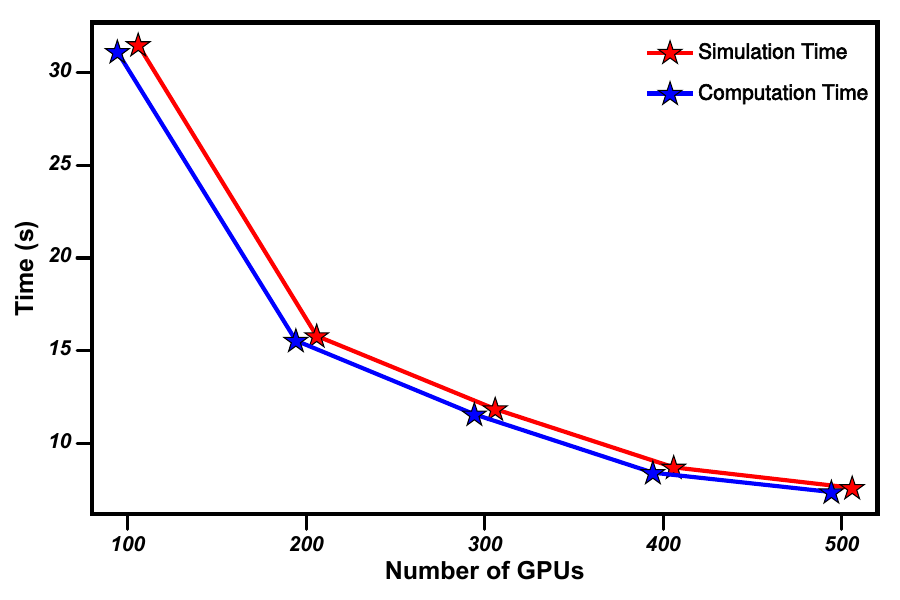}
    \caption{\textbf{Strong scaling property}. We controlled the number of neurons: 1.5 billion and varied the number of GPUs: 100, 200, 300, 400, 500. The horizontal axis is the number of GPUs, and the vertical axis is the time (s) spent for 1 sec simulation.}
    \label{fig:Ss}
\end{figure}

For the fixed scale of the cerebral cortex model with 1.5e6 neurons, the number of GPUs is changed from 100 to 1000, as shown in Fig.\ref{fig:Ss}. As the number of GPUs increases, the computation time cost continues to decrease, approximate inverse proportional function.

\subsection{Different firing rates}
As mentioned above, communication traffic transferred between nodes mainly consists of spike information. So the increase in the firing rates of the brain network leads to an increase in spikes, further leading to an increase in communication traffic. Meanwhile, the calculation of $J_{intra}$ and $J_{inter}$ sum the spikes of presynaptic neurons. Theoretically, as an increase in firing rate, the total computation time gradually decreases in a linear relationship. Here we vary the firing rate of the network from 10 to 160 by adjusting $g_{AMPA}$ in the equation\ref{eq:cv_equation}. Although the excessively high firing rates are biologically meaningless, we only aim to discuss the firing rate effect on brain simulation.

\begin{figure}[htbp]
    \centering
    \includegraphics[width=0.24\textwidth]{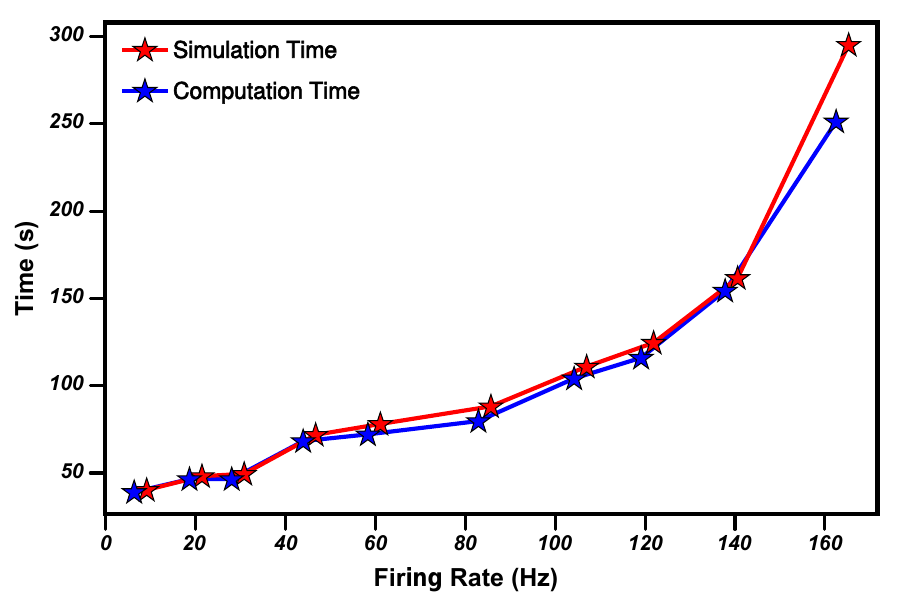}
    \caption{\textbf{Different firing rate}. The horizontal axis is the firing rate of the network, and the vertical axis is the time (s) spent for 1 sec simulation.}
    \label{fig:Fs}
\end{figure}

Firstly, we show the relationship between the total simulation time and different average firing rates as shown in Fig.\ref{fig:Fs}. It is evident that the higher the firing rate, the higher the total simulation time. When the firing rate is up to 169.9Hz, the total simulation time can reach up to 140s.

\begin{figure}[htbp]
    \centering
    \includegraphics[width=0.24\textwidth]{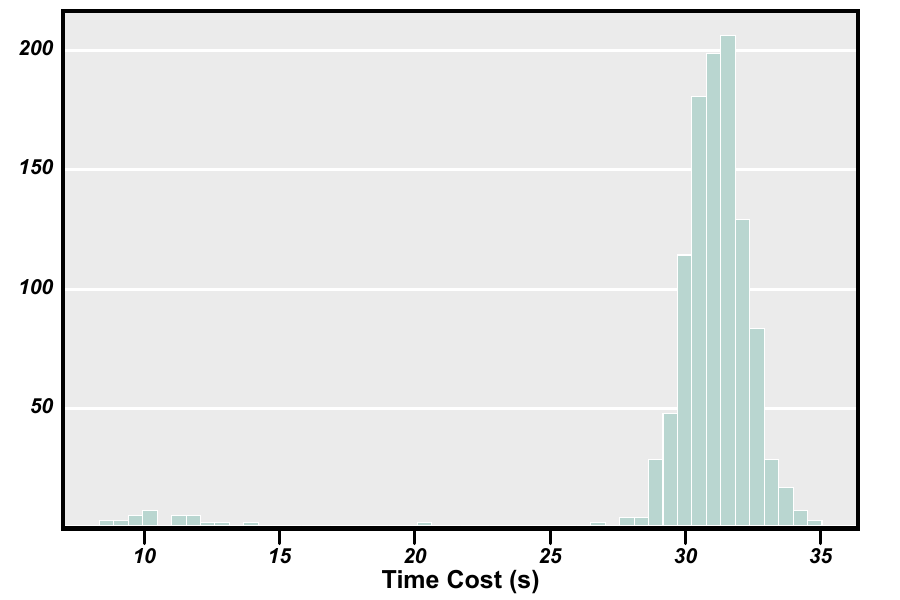}
    \includegraphics[width=0.24\textwidth]{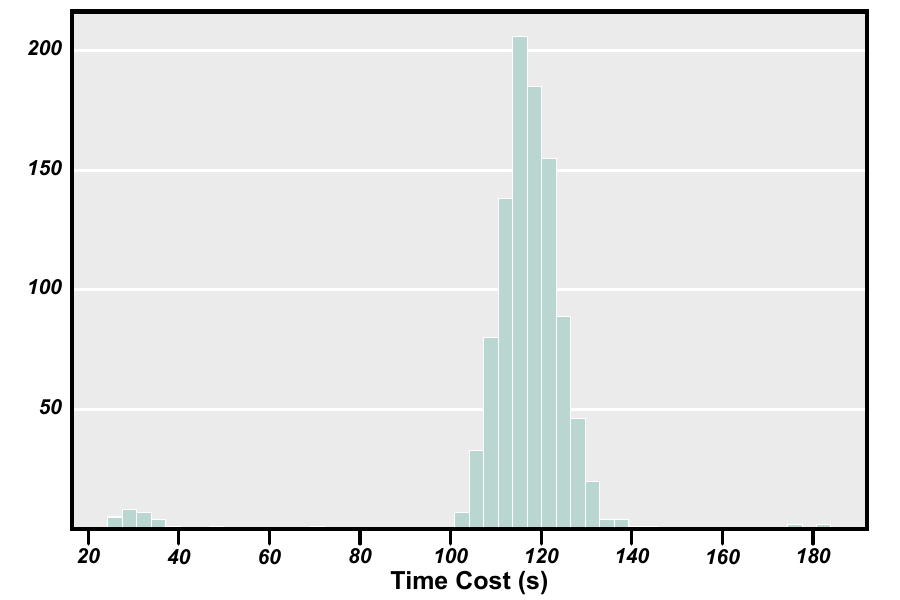}
    \caption{\textbf{The specific situation of computation time on GPUs with normal firing rate 7.7Hz(left) and the extreme firing rate 163.9Hz(right)}. The horizontal axis is the computation time (sec), and the vertical axis is the number of GPUs corresponding to the computation time interval. }
    \label{fig:normalandhigh}
\end{figure}

In order to further investigate the specific situation of computation time on GPUs, we have collected detailed computation times of threads on the different GPUs during the neuron equation update. We specifically compared results on the normal firing rate network (denoted as trial 1) and the highest firing rate network(denoted as trial 2) in Fig.\ref{fig:normalandhigh}. We define the time statistic $T_{max}$ and $T_{std}$ as follows and calculate the time statistic of two trials respectively.
\begin{equation*}
    T_{max,trial} = \frac{\mathop{max}\limits_{i\in \GPUs}T_{com, i}}{\mathop{mean}\limits_{i\in \GPUs} T_{com, i}},\quad
    T_{std,trial} = \frac{\mathop{std}\limits_{i\in \GPUs}T_{com, i}}{\mathop{mean}\limits_{i\in \GPUs} T_{com, i}}
\end{equation*}

Noticed $T_{max, 1} = 1.592 > T_{max, 2} = 1.143 $, $T_{std, 1} = 0.135 > T_{std, 2} = 0.110 $, it is easy to find that when the firing rate increases, the computation time distribution of different GPUs is more divergent and the ratio of maximum time to average time also increases. This indicates that most GPUs wait for the slowest GPU to end the computing process. From the perspective of parallel computing, the lack of a balanced computing load is not conducive to efficient simulation. Therefore, in the most ideal scenario, we should design the number of neurons loaded on different GPUs based on the brain firing rate, so that the whole network model can be load balanced and achieve higher simulation efficiency.

\section{Implication}
\subsection{Task simulation example on human-scale brain network}
\begin{figure}[ht]
 \centering
 \includegraphics[width=0.4\textwidth]{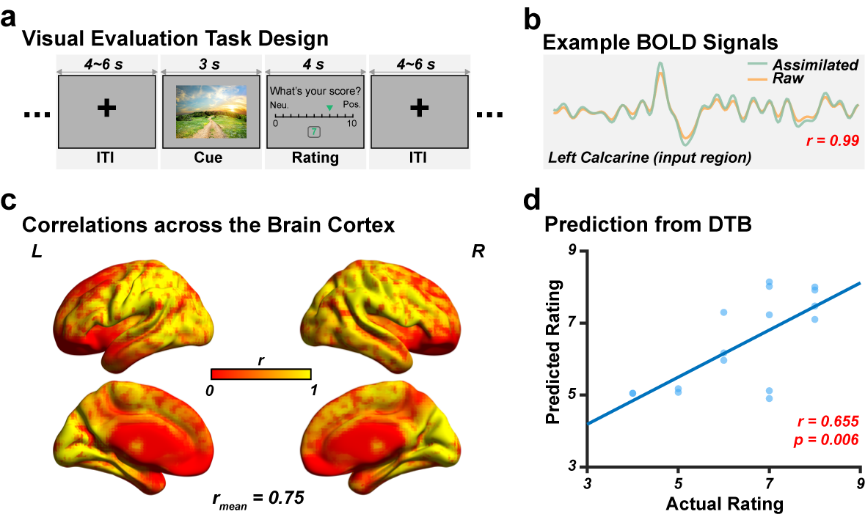}
 \caption{\textbf{Task assimilation with DTB}.
 a. A schematic illustration of the visual evaluation task. b. The illustration of the empirical and assimilated BOLD signals with a time lag of 2 in the left calcarine cortex (one of the input brain regions) during the visual evaluation task. c. The similarity between the assimilated digital task brain and the biological brain across the brain cortex is represented by the Pearson correlation coefficients at the voxel level. d. The predicted performance is based on the assimilated digital task brain of the visual evaluation task.
 }
 \label{fig:Task assimilation with DTB}
\end{figure}

To show the potential of DTB for applications, we mimicked the experimental BOLD signals in the task-related brain regions acquired from task functional MRI and utilized the assimilated hyperparameters of the perceptive input brain regions as the injection currents to obtain a digital task brain. Herein, a visual evaluation task was implemented in our DTB, which consisted of 18 positively emotional scenes. For each trial, after a 3-second figure cue, the participant was asked to evaluate how he feel about the visual stimulus with a Likert scale from 0 to 10 in 4 seconds (Fig.\ref{fig:Task assimilation with DTB}a). Three brain regions were included as the perceptive inputs: the bilateral calcarine cortices, which were referred as the primary visual cortex; the bilateral supramarginal gyrus, which was parts of the somatosensory network and the bilateral inferior frontal gyrus (opercular part), which could mediate maintenance of external stimulus information. The Pearson correlation coefficients between the assimilated BOLD signals and the corresponding real BOLD signals were computed as an estimation of the similarity between the digital task brain and the biological brain. The correlations could reach above 0.98 with a time lag of 2 in the perceptive input regions (Fig.\ref{fig:Task assimilation with DTB}b). More importantly, at the voxel level, the averaged similarity between the assimilated BOLD signals and the real BOLD signals across the brain cortex was 0.75 (Fig.\ref{fig:Task assimilation with DTB}c), indicating that the DTB in the visual evaluation task fitted well with its biological counterpart.

To further demonstrate the DTB performance, we used the assimilated digital task brain to predict the task scores in the real-world. To achieve this, the brain activation of the visual cue for each trial was first assessed with the BOLD signals from both the real and digital task brains by the general linear model, in which the regressors for modeling each trial were established by convolving the corresponding experimental condition with SPM canonical hemodynamic response function (HRF) and six head motion parameters were set as the additional covariate regressors. Hence, the patterns of brain activation were comparable between the digital brain and the biological brain. We then trained a linear regression model with the biological brain activations during the stimuli evaluation as the response variables and the real scores of the emotional pictures as the predictors. The least absolute shrinkage and selection operator (LASSO) regularization was employed to pursue a sparse coefficient vector. In this model, we removed the first two trials due to the bad quality of assimilation and excluded the brain activations in the input regions to minimize the confounding factors from the inputs. Finally, we predicted how the subject rated real-world pictures with the brain activations from the digital brain and the sparse coefficient vector obtained via LASSO. Remarkably, the predicted scores from the digital brain were consistent with the actual scores rated by the participants (r = 0.655, p = 0.006, Fig.\ref{fig:Task assimilation with DTB}d).

\subsection{Reverse engineering of brain function and brain-inspired intelligence}
The DTB, equipped with the capability to simulate human-brain-scaled spiking neuronal networks with biological structure and assimilation of the model up to trillions of parameters, is able to interact with the computing model in the real-world environment. Hence it could serve as a computing platform for digitally performing various cognitive and medical tasks, as demonstrated by the latest example, as well as possibly brain-machine interface experiments, and provide novel insights into the neurobehavioural mechanism of the human brain. Nevertheless, it could also conduct digital twin experiments of brain intelligence, in particular of human beings and thus establish a methodology of reverse engineering to investigate brain-inspired intelligence. The conflicts between the general HPC system and the brain computing system imply a research orientation to explore the new generation of architecture for brain-inspired computing systems.

\bibliographystyle{IEEEtran}
\bibliography{references}

\begin{thebibliography}{10}
\providecommand{\url}[1]{#1}
\csname url@samestyle\endcsname
\providecommand{\newblock}{\relax}
\providecommand{\bibinfo}[2]{#2}
\providecommand{\BIBentrySTDinterwordspacing}{\spaceskip=0pt\relax}
\providecommand{\BIBentryALTinterwordstretchfactor}{4}
\providecommand{\BIBentryALTinterwordspacing}{\spaceskip=\fontdimen2\font plus
\BIBentryALTinterwordstretchfactor\fontdimen3\font minus
  \fontdimen4\font\relax}
\providecommand{\BIBforeignlanguage}[2]{{%
\expandafter\ifx\csname l@#1\endcsname\relax
\typeout{** WARNING: IEEEtran.bst: No hyphenation pattern has been}%
\typeout{** loaded for the language `#1'. Using the pattern for}%
\typeout{** the default language instead.}%
\else
\language=\csname l@#1\endcsname
\fi
#2}}
\providecommand{\BIBdecl}{\relax}
\BIBdecl

\bibitem{Makin2019}
\BIBentryALTinterwordspacing
S.~Makin, ``The four biggest challenges in brain simulation,'' \emph{Nature},
  vol. 571, no. 7766, pp. S9--S9, Jul. 2019. [Online]. Available:
  \url{https://doi.org/10.1038/d41586-019-02209-z}
\BIBentrySTDinterwordspacing

\bibitem{HerculanoHouzel2009}
\BIBentryALTinterwordspacing
S.~Herculano-Houzel, ``The human brain in numbers: a linearly scaled-up primate
  brain,'' \emph{Frontiers in Human Neuroscience}, vol.~3, 2009. [Online].
  Available: \url{https://doi.org/10.3389/neuro.09.031.2009}
\BIBentrySTDinterwordspacing

\bibitem{Markram2006}
\BIBentryALTinterwordspacing
H.~Markram, ``The blue brain project,'' \emph{Nature Reviews Neuroscience},
  vol.~7, no.~2, pp. 153--160, Feb. 2006. [Online]. Available:
  \url{https://doi.org/10.1038/nrn1848}
\BIBentrySTDinterwordspacing

\bibitem{Amunts2016}
\BIBentryALTinterwordspacing
K.~Amunts, C.~Ebell \emph{et~al.}, ``The human brain project: Creating a
  european research infrastructure to decode the human brain,'' \emph{Neuron},
  vol.~92, no.~3, pp. 574--581, Nov. 2016. [Online]. Available:
  \url{https://doi.org/10.1016/j.neuron.2016.10.046}
\BIBentrySTDinterwordspacing

\bibitem{SanzLeon2013}
\BIBentryALTinterwordspacing
P.~S. Leon, S.~A. Knock \emph{et~al.}, ``The virtual brain: a simulator of
  primate brain network dynamics,'' \emph{Frontiers in Neuroinformatics},
  vol.~7, 2013. [Online]. Available:
  \url{https://doi.org/10.3389/fninf.2013.00010}
\BIBentrySTDinterwordspacing

\bibitem{https://doi.org/10.48550/arxiv.2211.15963}
\BIBentryALTinterwordspacing
W.~Lu, Q.~Zheng \emph{et~al.}, ``The human digital twin brain in the resting
  state and in action,'' 2022. [Online]. Available:
  \url{https://arxiv.org/abs/2211.15963}
\BIBentrySTDinterwordspacing

\bibitem{Binzegger2009}
\BIBentryALTinterwordspacing
T.~Binzegger, R.~Douglas, and K.~Martin, ``Topology and dynamics of the
  canonical circuit of cat v1,'' \emph{Neural Networks}, vol.~22, no.~8, pp.
  1071--1078, Oct. 2009. [Online]. Available:
  \url{https://doi.org/10.1016/j.neunet.2009.07.011}
\BIBentrySTDinterwordspacing

\bibitem{Finnema2016}
\BIBentryALTinterwordspacing
S.~J. Finnema, N.~B. Nabulsi \emph{et~al.}, ``Imaging synaptic density in the
  living human brain,'' \emph{Science Translational Medicine}, vol.~8, no. 348,
  Jul. 2016. [Online]. Available:
  \url{https://doi.org/10.1126/scitranslmed.aaf6667}
\BIBentrySTDinterwordspacing

\bibitem{Hansen2022}
\BIBentryALTinterwordspacing
J.~Y. Hansen, G.~Shafiei \emph{et~al.}, ``Local molecular and global
  connectomic contributions to cross-disorder cortical abnormalities,''
  \emph{Nature Communications}, vol.~13, no.~1, Aug. 2022. [Online]. Available:
  \url{https://doi.org/10.1038/s41467-022-32420-y}
\BIBentrySTDinterwordspacing

\bibitem{Farley1954}
\BIBentryALTinterwordspacing
B.~Farley and W.~Clark, ``Simulation of self-organizing systems by digital
  computer,'' \emph{Transactions of the {IRE} Professional Group on Information
  Theory}, vol.~4, no.~4, pp. 76--84, Sep. 1954. [Online]. Available:
  \url{https://doi.org/10.1109/tit.1954.1057468}
\BIBentrySTDinterwordspacing

\bibitem{Rochester1956}
\BIBentryALTinterwordspacing
N.~Rochester, J.~Holland \emph{et~al.}, ``Tests on a cell assembly theory of
  the action of the brain, using a large digital computer,'' \emph{{IEEE}
  Transactions on Information Theory}, vol.~2, no.~3, pp. 80--93, Sep. 1956.
  [Online]. Available: \url{https://doi.org/10.1109/tit.1956.1056810}
\BIBentrySTDinterwordspacing

\bibitem{Brette2007}
\BIBentryALTinterwordspacing
R.~Brette, M.~Rudolph \emph{et~al.}, ``Simulation of networks of spiking
  neurons: A review of tools and strategies,'' \emph{Journal of Computational
  Neuroscience}, vol.~23, no.~3, pp. 349--398, Jul. 2007. [Online]. Available:
  \url{https://doi.org/10.1007/s10827-007-0038-6}
\BIBentrySTDinterwordspacing

\bibitem{Gara2005}
\BIBentryALTinterwordspacing
A.~Gara, M.~A. Blumrich \emph{et~al.}, ``Overview of the blue gene/l system
  architecture,'' \emph{{IBM} Journal of Research and Development}, vol.~49,
  no. 2.3, pp. 195--212, Mar. 2005. [Online]. Available:
  \url{https://doi.org/10.1147/rd.492.0195}
\BIBentrySTDinterwordspacing

\bibitem{Frye2007IBMRR}
J.~Frye, R.~Ananthanarayanan, and D.~S. Modha, ``Ibm research report towards
  real-time, mouse-scale cortical simulations,'' 2007.

\bibitem{Ananthanarayanan2007b}
\BIBentryALTinterwordspacing
R.~Ananthanarayanan and D.~S. Modha, ``Scaling, stability and synchronization
  in mouse-sized (and larger) cortical simulations,'' \emph{{BMC}
  Neuroscience}, vol.~8, no.~S2, Jul. 2007. [Online]. Available:
  \url{https://doi.org/10.1186/1471-2202-8-s2-p187}
\BIBentrySTDinterwordspacing

\bibitem{Ananthanarayanan2007}
\BIBentryALTinterwordspacing
------, ``Anatomy of a cortical simulator,'' in \emph{Proceedings of the 2007
  {ACM}/{IEEE} conference on Supercomputing}.\hskip 1em plus 0.5em minus
  0.4em\relax {ACM}, Nov. 2007. [Online]. Available:
  \url{https://doi.org/10.1145/1362622.1362627}
\BIBentrySTDinterwordspacing

\bibitem{Izhikevich2008}
\BIBentryALTinterwordspacing
E.~M. Izhikevich and G.~M. Edelman, ``Large-scale model of mammalian
  thalamocortical systems,'' \emph{Proceedings of the National Academy of
  Sciences}, vol. 105, no.~9, pp. 3593--3598, Mar. 2008. [Online]. Available:
  \url{https://doi.org/10.1073/pnas.0712231105}
\BIBentrySTDinterwordspacing

\bibitem{Ananthanarayanan2009}
\BIBentryALTinterwordspacing
R.~Ananthanarayanan, S.~K. Esser \emph{et~al.}, ``The cat is out of the bag,''
  in \emph{Proceedings of the Conference on High Performance Computing
  Networking, Storage and Analysis}.\hskip 1em plus 0.5em minus 0.4em\relax
  {ACM}, Nov. 2009. [Online]. Available:
  \url{https://doi.org/10.1145/1654059.1654124}
\BIBentrySTDinterwordspacing

\bibitem{Gewaltig2007}
\BIBentryALTinterwordspacing
M.-O. Gewaltig and M.~Diesmann, ``{NEST} ({NEural} simulation tool),''
  \emph{Scholarpedia}, vol.~2, no.~4, p. 1430, 2007. [Online]. Available:
  \url{https://doi.org/10.4249/scholarpedia.1430}
\BIBentrySTDinterwordspacing

\bibitem{Helias2012}
\BIBentryALTinterwordspacing
M.~Helias, S.~Kunkel \emph{et~al.}, ``Supercomputers ready for use as discovery
  machines for neuroscience,'' \emph{Frontiers in Neuroinformatics}, vol.~6,
  2012. [Online]. Available: \url{https://doi.org/10.3389/fninf.2012.00026}
\BIBentrySTDinterwordspacing

\bibitem{Ajima2009TofuA6}
Y.~Ajima, S.~Sumimoto, and T.~Shimizu, ``Tofu: A 6d mesh/torus interconnect for
  exascale computers,'' \emph{Computer}, vol.~42, 2009.

\bibitem{Kunkel2014}
\BIBentryALTinterwordspacing
S.~Kunkel, M.~Schmidt \emph{et~al.}, ``Spiking network simulation code for
  petascale computers,'' \emph{Frontiers in Neuroinformatics}, vol.~8, Oct.
  2014. [Online]. Available: \url{https://doi.org/10.3389/fninf.2014.00078}
\BIBentrySTDinterwordspacing

\bibitem{Igarashi2019}
\BIBentryALTinterwordspacing
J.~Igarashi, H.~Yamaura, and T.~Yamazaki, ``Large-scale simulation of a layered
  cortical sheet of spiking network model using a tile partitioning method,''
  \emph{Frontiers in Neuroinformatics}, vol.~13, Nov. 2019. [Online].
  Available: \url{https://doi.org/10.3389/fninf.2019.00071}
\BIBentrySTDinterwordspacing

\bibitem{Gosui2016}
\BIBentryALTinterwordspacing
M.~Gosui and T.~Yamazaki, ``Real-world-time simulation of memory consolidation
  in a large-scale cerebellar model,'' \emph{Frontiers in Neuroanatomy},
  vol.~10, Mar. 2016. [Online]. Available:
  \url{https://doi.org/10.3389/fnana.2016.00021}
\BIBentrySTDinterwordspacing

\bibitem{Yamazaki2017}
\BIBentryALTinterwordspacing
T.~Yamazaki, J.~Igarashi \emph{et~al.}, ``Real-time simulation of a cat-scale
  artificial cerebellum on {PEZY}-{SC} processors,'' \emph{The International
  Journal of High Performance Computing Applications}, vol.~33, no.~1, pp.
  155--168, Jun. 2017. [Online]. Available:
  \url{https://doi.org/10.1177/1094342017710705}
\BIBentrySTDinterwordspacing

\bibitem{Yamaura2020}
\BIBentryALTinterwordspacing
H.~Yamaura, J.~Igarashi, and T.~Yamazaki, ``Simulation of a human-scale
  cerebellar network model on the k computer,'' \emph{Frontiers in
  Neuroinformatics}, vol.~14, Apr. 2020. [Online]. Available:
  \url{https://doi.org/10.3389/fninf.2020.00016}
\BIBentrySTDinterwordspacing

\bibitem{6468524}
R.~Preissl, T.~M. Wong \emph{et~al.}, ``Compass: A scalable simulator for an
  architecture for cognitive computing,'' in \emph{SC '12: Proceedings of the
  International Conference on High Performance Computing, Networking, Storage
  and Analysis}, 2012, pp. 1--11.

\bibitem{icpp2022}
Y.~Liu, X.~Du \emph{et~al.}, ``Regularizing sparse and imbalanced
  communications for voxel-based brain simulations on supercomputers,'' in
  \emph{Proceedings of the 51st International Conference on Parallel
  Processing}, 2022, pp. 1--10.

\bibitem{9785756}
X.~Du, Y.~Liu \emph{et~al.}, ``A low-latency communication design for brain
  simulations,'' \emph{IEEE Network}, vol.~36, no.~2, pp. 8--15, 2022.

\bibitem{Friston2000}
\BIBentryALTinterwordspacing
K.~Friston, A.~Mechelli \emph{et~al.}, ``Nonlinear responses in fmri: The
  balloon model, volterra kernels, and other hemodynamics,'' \emph{NeuroImage},
  vol.~12, pp. 466--477, 10 2000. [Online]. Available:
  \url{https://linkinghub.elsevier.com/retrieve/pii/S105381190090630X}
\BIBentrySTDinterwordspacing

\bibitem{https://doi.org/10.48550/arxiv.2206.02986}
\BIBentryALTinterwordspacing
W.~Zhang, B.~Chen \emph{et~al.}, ``On a framework of data assimilation for
  neuronal networks,'' 2022. [Online]. Available:
  \url{https://arxiv.org/abs/2206.02986}
\BIBentrySTDinterwordspacing

\end{thebibliography}
\end{document}